\documentclass[runningheads]{llncs}
\usepackage{graphicx}
\usepackage{amsmath}
\usepackage{threeparttable}

\begin{document}
\title{A Hybrid 3DCNN and 3DC-LSTM based model for 4D Spatio-temporal fMRI data: An ABIDE Autism Classification study}
\titlerunning{Hybrid 3D CNNs and 3D C-LSTMs }
%
\author{Ahmed El-Gazzar\inst{1} \and
Mirjam Quaak\inst{1,2} \and
Leonardo Cerliani\inst{1} \and
Peter Bloem\inst{2} \and
Guido van Wingen\inst{1} \and
Rajat Mani Thomas\inst{1}}
\authorrunning{A. El-Gazzar et al.}
%
\institute{{Amsterdam UMC, University of Amsterdam, \\
Department of Psychiatry, Amsterdam, The Netherlands \\} 
\and
{Vrije Universiteit Amsterdam, Amsterdam, The Netherlands}}
\maketitle              

\begin{abstract}
Functional Magnetic Resonance Imaging (fMRI) captures the temporal dynamics of neural activity as a function of spatial location in the brain. Thus, fMRI scans are represented as 4-Dimensional (3-space + 1-time) tensors. And it is widely believed that the spatio-temporal patterns in fMRI manifests as behaviour and clinical symptoms. Because of the high dimensionality ($\sim$ 1 Million) of fMRI, and the added constraints of limited cardinality of data sets, extracting such patterns are challenging. A standard approach to overcome these hurdles is to  reduce the dimensionality of the data by either summarizing activation over time or space at the expense of possible loss of useful information. Here, we introduce an end-to-end algorithm capable of extracting spatiotemporal features from the full 4-D data using 3-D CNNs and 3-D Convolutional LSTMs. We evaluate our proposed model on the publicly available ABIDE dataset to demonstrate the capability of our model to classify Autism Spectrum Disorder (ASD) from resting-state fMRI data. Our results show that the proposed model achieves state of the art results on single sites with F1-scores of 0.78 and 0.7 on NYU and UM sites, respectively.

\end{abstract}
\section{Introduction}
Unlike other fields of medicine, psychiatry lacks diagnostic criteria based on validated biomarkers. Finding these biomarkers is critical for (i) understanding the underlying neural causes, (ii) improving diagnosis and (iii) predicting treatment outcome. Functional magnetic resonance imaging (fMRI) --- a well-established proxy for neural activity --- is often taunted as a promising non-invasive technique that has enough information in them to design a robust biomarker. This information, often present as spatio-temporal patterns in fMRI is challenging to extract given its dimensionality ($\sim$ 1 Million) and typical data volumes (typically $< 200$ samples/subjects at any given center). In this paper we focus on Autism Spectrum Disorder (ASD). ASD represents a heterogeneous group of developmental brain disorders characterized by lifelong social deficits and repetitive behaviour. \\
Deep learning, because of its recent success in a multitude of tasks, is being currently explored in neuroimaging. For example in classifying Alzheimer, and predicting disease conversion \cite{vieira2017using}. The key advantage of deep learning is its ability to learn useful features from raw data; eliminating the need for subjective feature design as required by "classical" machine learning techniques. But applying deep learning to fMRI has been problematic because of the issue of dimensionality and data volume.\\
To overcome these issues fMRI data are often reduced in dimension either by summarizing brain activity spatially or temporally. In the classification of ASD versus controls for example, several studies convert the full 4-D resting-state fMRI (rs-fMRI) signal in to a correlation matrix. These matrices are based on the average time course within regions-of-interest (ROI) given by an atlas  \cite{heinsfeld2018identification,abraham2017deriving,khosla20183d}. Instead of averaging over time, Dvornek et al. \cite{dvornek2017identifying} used long short-term memory (LSTM) cells on the timeseries of 200 selected brain regions for the same task. Similarly, \cite{1d} applied 1D convolutions on extracted timeseries of different atlases. Alternatively, Li et al. \cite{li20182} directly learned spatial features from the 3D fMRI images, but reduced the temporal information by taking the mean and standard deviation of fixed time windows. These subjective feature selection methods  could drastically reduce the ability to detect complex patterns in neural activity and may lead to suboptimal results.\\
Instead, we propose to learn end-to-end from the full 4D fMRI sequences using a framework that takes advantage of both spatial and temporal information in the data to achieve the objective. On the ABIDE dataset, we show that our approach can surpass subjective methods that rely on feature engineering and we also avoid any procedure to summarize data. 

\begin{figure}[!ht]
\includegraphics[width=\textwidth]{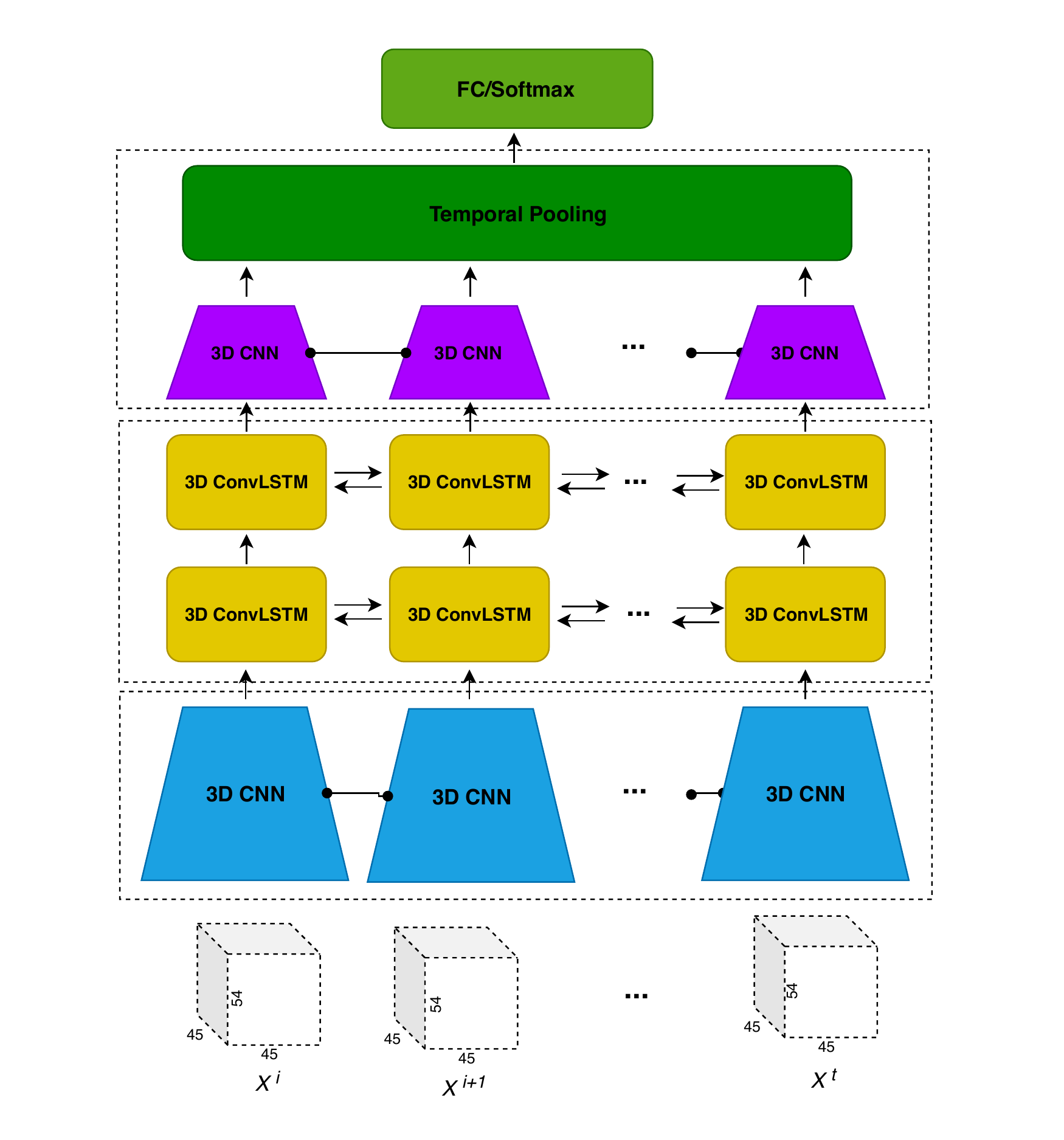}
\caption{Overview of the proposed deep architecture. 3DCNN
and bidirectional C-LSTM are used to learn the spatial and long-term spatiotemporal features, following which
a 3DCNN is used to learn higher-level spatiotemporal features based
on the learnt 3D long-term spatiotemporal feature maps for the final classification layer.} \label{fig1}
\end{figure}

\section{Method}
In this work we present a novel architecture for 4D rs-fMRI data with application to ASD classification. We utilize the strength of convolutional LSTMs (C-LSTM) in spatio-temporal feature extraction by employing a 3D variant in our proposed pipeline. Further, we demonstrate another variation of architecture that does not use convolutional LSTMs. Since LSTMs are computationally expensive, we propose a computationally cheaper alternative with a 1D convolution for spatio-temporal processing. The idea of this variant was inspired by \cite{1d} that demonstrated the capability of 1D convolutions to extract useful features from time courses of rs-fMRI for the diagnosis of ASD. In addition, the comparison of these two models helps to identify the contribution of the convolutional LSTMs.  


\subsection{3DCNN\_C-LSTM}
The main challenge in employing recurrent neural networks (RNN) in fMRI is the high dimensionality of the data. The spatial dimension of a 4mm down-sampled volume in MNI space is 45x54x45, where the size of the time-series depends on the duration of the scan and the TR, and usually ranges from 100 to 400 time points. Together with the limited sample sizes, classical RNNs fail to train efficiently on the raw 4D volumes. One alternative is to reduce the spatial dimensions first, but that is likely to remove informative local/temporal features. To overcome this issue, we design an end-to-end pipeline that enables efficient training of RNNs in high dimensional environments.
Our pipeline consists of three components:

\paragraph{3D CNN for spatial feature learning:}
The 3D CNN component in the proposed architecture is a shallow CNN with 4 convolutional layers. The purpose of this component is: (1) to reduce the spatial dimension of the original volume for efficient training of the recurrent layers.
(2) extract lower level spatial feature maps necessary for spatio-temporal feature learning at the next stage. We use a 3D CNN with tied weights at all the input time steps for coherent spatial feature extraction and efficient training. \\
The kernel size of each CNN layer is $3 \times  3 \times 3$ with stride $2 \times 2 \times 2$ to down-sample the input feature vector. We add dropout with a rate of 0.2 to the output of every convolution to regularize the network.

\paragraph{3D C-LSTM for spatio-temporal feature learning:}
One of the most common choices to model temporal sequences is the LSTM. Unfortunately, LSTMs take a sequence of vectors as inputs. This would require us to flatten our spatial dimensions, and thus ignore spatial patterns. Moreover, the LSTM applies fully connected transformations to these vectors, leading to very large weight matrices, unless the spatial dimensions are strongly reduced. The C-LSTM \cite{xingjian2015convolutional} solves both problems: it replaces the fully connected vector-transformations by convolutions, allowing us to model the temporal information in a memory efficient way, without flattening the spatial dimensions. \\

 The inputs $X_1, ..., X_t$ , the cell states $C_1, ..., C_t$
, the hidden states $H_1, ..., H_t$ and the gates $i_t$, $f_t$, $o_t$ of C-LSTM are all 4D tensors. Let $\ast$ denote the convolution operator, and let $\otimes$ denote the Hadamard product. The C-LSTM can be formulated as:

 
 



\begin{align}
\begin{split}\label{eq:1}
  i_{t} = \sigma (W_{xi} \ast X_{t} + W_{hi} \ast H_{t-1} + b_{i} )
\end{split}\\
\begin{split}\label{eq:2}
    f_{t} = \sigma (W_{xf}  \ast X_{t} + W_{hf} \ast H_{t-1} + b_{f})
\end{split}\\
\begin{split}\label{eq:3}
   o_{t} = \sigma (W_{xo}  \ast X_{t} + W_{ho} \ast H_{t-1} + b_{o})
\end{split}\\
\begin{split}\label{eq:4}
   C_{t} = f_{t} \otimes C_{t-1} + i_{t} \text{tanh} (W_{xc} \ast X_{t} + W_{hc} \ast H_{t-1} + b_{c})
\end{split}\\
\begin{split}\label{eq:5}
H_{t} = o_t \otimes \text{tanh}(C_t)
\end{split}
\end{align}

Where $\sigma$ is the sigmoid function, and all weight matrices $W$ are 3D convolution kernels.
The convolutions in the C-LSTM have kernel size $3 \times
3$ with stride $1 \times 1$. ”Same-Padding” is used to ensure
that the spatiotemporal feature maps in each C-LSTM
layer have the same spatial size. A two-layer bidirectional
C-LSTM is constructed as illustrated in Figure \ref{fig1} to encode global temporal information and local spatial information into 3D saptio-temporal feature maps.

\paragraph{3D CNN for higher level spatio-temporal feature learning:}
Since the 3D spatiotemporal feature maps still have large
spatial size, dimensionality reduction is necessary for the
final classification. Another simple 3DCNN with tied weights is employed  to reduce the dimensionality further and to learn the higher-level spatiotemporal features, based on the learnt 3D spatiotemporal feature maps at each recurrent step of C-LSTM.
Only a shallow 3DCNN is constructed in this implementation. Nevertheless, deeper 3DCNNs can also be used for different configurations or
applications.

\subsection{3DCNN\_1D}
1D convolutions offer a simpler alternative to LSTMs with longer effective memory \cite{bai2018empirical}. They have been successfully applied to capture the temporal dynamics of the fMRI signal for ASD classification \cite{1d}. Therefore, this alternative model applies a 1D convolution for spatio-temporal feature learning after the 3D CNN component. The first layers for spatial feature learning are similar to the 3DCNN\_C-LSTM model.  After the 3D convolutional layers,  a global average pooling layer is added to yield a 1D vector with the length of the input time-series. One 1D Convolution is applied on this vector with the learned spatial features as input channels. Hereafter, a temporal pooling layer as in the C-LSTM model is conducted to summarize the temporal information followed by a fully connected layer to output the classification probabilities.

\section{Experiments and Results}

\subsection{Datasets}
We use the publicly available ABIDE dataset to evaluate our proposed pipeline. We preprocessed the data with the Configurable Pipeline for the Analysis of Connectomes (C-PAC)  and the fMRI volumes are downsampled to 4 mm in MNI. We use single sites to evaluate the network capacity to learn the spatio-temporal features with a small sample size but uniform scanning parameters. We also experiment with the multi-site data provided from ABIDE-I dataset to test the network performance in a heterogeneous environment but a larger sample size. For single sites experiments, we use the NYU and UM sites from ABIDE-I. Those provide the  highest number of balanced (ASD/typically developing (TD)) subjects with 184 and 110 subjects respectively.
For the multi-site experiment we used ABIDE-I with 19 sites and 1100 subjects.

\subsection{Network Training}
The proposed architectures are trained in an end-to-end fashion from scratch. To speed up training and to increase the diversity of samples seen by the model, we select a random contiguous sub-sequence of 20 time points for each instance (re-sampling every epoch). For validation and testing, the full time-series per subject are used by feeding subsequent crops to the model and average the predictions over all crops. We train our models for 500 epochs with a batch size of 8. For
optimization of the cross entropy loss function, we employ the Adam optimizer with a learning rate of 0.0001. During training, we evaluate the performance on the validation set every 10 epochs and use the best model for evaluation.

\subsection{Results and comparison with state-of-the-art}
We compare our models to previous deep learning ABIDE classification models that handled the temporal and spatial dimensions in different ways and achieved the best results reported thus far. We report the results for the models in Table \ref{tab:results} on the respective dataset. We reran the best reported experiments from \cite{heinsfeld2018identification} and  using the recommended settings and available code on our dataset and report the results. For \cite{1d} we used the full time-series for single-site experiments and selected for and cropped to 100 timepoints for the ABIDE-I dataset. For \cite{bengs20194d} we report the results for the models on the NYU site from their paper. We provide a short description of the models and input data type. 
\begin{itemize}
    \item \textbf{AE\_MLP \cite{heinsfeld2018identification}{}:} uses correlation matrices of the extracted time-courses from the Craddock atlas \cite{craddock2012whole} to pre-train a stacked fully-connected autoencoder and fine tune it for classification.
    \item \textbf{SVM \cite{svm}:} uses the same input features as \textbf{AE\_MLP} to train a support vector machine with an rbf kernel. 
    \item \textbf{1DConv \cite{1d}:} uses extracted time courses from the Harvard Oxford atlas as input to a 1D convolutional neural network.
    \item \textbf{CNN3D\_TC \cite{bengs20194d}:} 3D spatial data is used in a 3D convolutional network where the temporal information is stacked as channels.
    \item \textbf{CNN3D\_MD \cite{bengs20194d}:} same approach as CNN\_TC but only mean and standard deviation of the temporal dimension are stacked as channels.
    \item \textbf{convGRU\_CNN3D \cite{bengs20194d}:} uses the 4D volume where spatio-temporal information are processed by a 3D convolutional GRU followed by a 3D CNN.
    \item \textbf{CNN4D \cite{bengs20194d}:} uses 4D convolutions on the the 4D volumetric data.
\end{itemize}

\begin{table}[]
\centering
\caption{5-fold cross validation mean accuracies and F1-scores of trained models on NYU, UM and ABIDE-I data}
\label{tab:results}
\begin{tabular}{lllll}
\textbf{Data} & \textbf{Model} & \textbf{Accuracy} &   \textbf{F1-score}  \\ \hline

 NYU             & AE\_MLP \cite{heinsfeld2018identification}        &     $0.64\pm0.1$       &      \hspace{3mm}$0.67$           \\
              & SVM \cite{svm}        &    $0.6\pm0.13$       &      \hspace{3mm}$0.59$            \\
              & 1D\_Conv  \cite{1d}     &     $0.64\pm0.11$       &     \hspace{3mm}$0.62$              \\
              & CNN3D\_TC\text{*} \cite{bengs20194d}   &     $0.57$          &  \hspace{3mm}$0.61$         \\
              & CNN3D\_MS\text{*}  \cite{bengs20194d}     &     $0.60$          &  \hspace{3mm}$0.65$             \\
              & convGRU-CNN3D\text{*} \cite{bengs20194d} &     $0.67$          &  \hspace{3mm}$0.71$                 \\
              & CNN4D\text{*} \cite{bengs20194d}        &     $0.60$          &   \hspace{3mm}$0.68$               \\
         & 3DCNN\_1D (ours)      &     $0.59\pm0.07$          &     \hspace{3mm}$0.58$              \\
              & \textbf{3DCNN\_C-LSTM (ours)}  &   \textbf{0.77$\pm$0.05}                &    \hspace{3mm}\textbf{0.78}               \\
              \hline

UM              & AE\_MLP \cite{heinsfeld2018identification}        &     $0.56\pm0.11$      &       \hspace{3mm}$0.59$          \\
              & SVM \cite{svm}        &     $0.54\pm0.11$       &      \hspace{3mm}$0.56$            \\
              & 1D\_Conv \cite{1d}    &     $0.63\pm0.1$         &       \hspace{3mm}$0.62$      \\
              & 3DCNN\_1D (ours)      &     $0.66\pm0.09$          &       \hspace{3mm}0.58        \\
              & \textbf{3DCNN\_C-LSTM (ours)}    &      \textbf{0.71$\pm$0.06}             &  \hspace{3mm}\textbf{0.70}                 \\ \hline
 
ABIDE-I              & AE\_MLP \cite{heinsfeld2018identification}       &      $0.63\pm0.02$         &        \hspace{3mm}$0.64$       \\
              & SVM \cite{svm}        &     $0.58\pm0.04$       &      \hspace{3mm}$0.6$            \\
              & \textbf{1D\_Conv}   \cite{1d}   &      \textbf{0.64$\pm$0.06}         &        \hspace{3mm}\textbf{0.64}          \\
              & 3DCNN\_1D (ours)      &   $0.54\pm0.02$        &        \hspace{3mm}$0.50$           \\ 
              & 3DCNN\_C-LSTM  (ours)     &    $0.58\pm0.03$         &         \hspace{3mm}$0.53$       \\ \hline
\end{tabular}
     \begin{tablenotes}
        \item  \text{*} Results as reported by Bengs et al. \cite{bengs20194d} on NYU data.
     \end{tablenotes}
\end{table}

\noindent We report 5-fold cross validation mean  F1-score and accuracy for the experiments in Table \ref{tab:results}. The results show that the proposed architecture 3DCNN\_C-LSTM outperforms other models on single site experiments by achieving  mean test accuracies and F1 scores of 0.77 and 0.78 respectively for the NYU site and 0.71 and 0.7 on the UM site. This surpasses previous methods by 10$\%$ and 8$\%$ for NYU and UM sites respectively.

3DCNN\_C-LSTM however also shows a degraded performance in multi-site environment as evidenced by the results on ABIDE-1 data that features 19 sites. We attribute the loss of performance of our model to the heterogeneity of the data acquired from different scanners with different scanning parameters. This effect does not show in other methods that do not use the full 4D volumes where data preprocessing and summarization play an important role in input signal consistency and hence model generalization.

Our results for the 3DCNN\_1D shows inferior performance compared to using C-LSTMs in all three datasets. This supports the vital role of a recurrent module in the network for spatio-temporal feature processing. However, the competitive performance of this architecture with the 1DConv model shows the ability of our first 3DCNN to extract useful spatial features in an end-to-end fashion compared to using pre-computed atlases.

\section{Discussion}
We have introduced a deep architecture that extracts information from fMRI signals for the classification of ASD, using 3DCNN and bidirectional 3DC-LSTMs; allowing the network to exploit local and global spatio-temporal structures. The proposed deep architecture provides an alternative method to hard-coded features or summary measures to reduce the dimensionality. The paper only presents the preliminary version of the deep architecture. The 3DCNN and C-LSTM networks can be further improved in order to obtain higher classification accuracy.
This architecture can also be used as a starting point for domain adaption techniques that can be deployed to boost the performance on multi-site data by compensating for data heterogeneity when using the full 4D volumes.

\section*{Acknowledgement}
This work was supported by the Netherlands Organization for Scientific Research (NWO; 628.011.023), Philips Research, AAA Data Science Program, and ZonMW (Vidi; 016.156.318).

%
%
%
%

\nocite{*} 
\bibliographystyle{splncs04}
\bibliography{paper007.bib}

\begin{thebibliography}{10}
\providecommand{\url}[1]{\texttt{#1}}
\providecommand{\urlprefix}{URL }
\providecommand{\doi}[1]{https://doi.org/#1}

\bibitem{abraham2017deriving}
Abraham, A., Milham, M.P., Di~Martino, A., Craddock, R.C., Samaras, D.,
  Thirion, B., Varoquaux, G.: Deriving reproducible biomarkers from multi-site
  resting-state data: An autism-based example. NeuroImage  \textbf{147},
  736--745 (2017)

\bibitem{anirudh2017bootstrapping}
Anirudh, R., Thiagarajan, J.J.: Bootstrapping graph convolutional neural
  networks for autism spectrum disorder classification. arXiv preprint
  arXiv:1704.07487  (2017)

\bibitem{bai2018empirical}
Bai, S., Kolter, J.Z., Koltun, V.: An empirical evaluation of generic
  convolutional and recurrent networks for sequence modeling. arXiv preprint
  arXiv:1803.01271  (2018)

\bibitem{belmonte2004autism}
Belmonte, M.K., Allen, G., Beckel-Mitchener, A., Boulanger, L.M., Carper, R.A.,
  Webb, S.J.: Autism and abnormal development of brain connectivity. Journal of
  Neuroscience  \textbf{24}(42),  9228--9231 (2004)

\bibitem{bengs20194d}
Bengs, M., Gessert, N., Schlaefer, A.: 4d spatio-temporal deep learning with 4d
  fmri data for autism spectrum disorder classification. Extended Abstract MIDL
  submission from https://openreview.net/forum?id=HklAUVnV5V  (2019)

\bibitem{craddock2012whole}
Craddock, R.C., James, G.A., Holtzheimer~III, P.E., Hu, X.P., Mayberg, H.S.: A
  whole brain fmri atlas generated via spatially constrained spectral
  clustering. Human brain mapping  \textbf{33}(8),  1914--1928 (2012)

\bibitem{svm}
Cristianini, N., Shawe-Taylor, J., et~al.: An introduction to support vector
  machines and other kernel-based learning methods. Cambridge university press
  (2000)

\bibitem{dvornek2017identifying}
Dvornek, N.C., Ventola, P., Pelphrey, K.A., Duncan, J.S.: Identifying autism
  from resting-state fmri using long short-term memory networks. In:
  International Workshop on Machine Learning in Medical Imaging. pp. 362--370.
  Springer (2017)

\bibitem{1d}
El~Gazzar, A., Cerliani, L., van Wingen, G., Mani~Thomas, R.: Simple 1-d
  convolutional networks for resting-state fmri based classification in autism
  (07 2019)

\bibitem{heinsfeld2018identification}
Heinsfeld, A.S., Franco, A.R., Craddock, R.C., Buchweitz, A., Meneguzzi, F.:
  Identification of autism spectrum disorder using deep learning and the abide
  dataset. NeuroImage: Clinical  \textbf{17},  16--23 (2018)

\bibitem{khosla20183d}
Khosla, M., Jamison, K., Kuceyeski, A., Sabuncu, M.: 3d convolutional neural
  networks for classification of functional connectomes. arXiv preprint
  arXiv:1806.04209  (2018)

\bibitem{li20182}
Li, X., Dvornek, N.C., Papademetris, X., Zhuang, J., Staib, L.H., Ventola, P.,
  Duncan, J.S.: 2-channel convolutional 3d deep neural network (2cc3d) for fmri
  analysis: Asd classification and feature learning. In: 2018 IEEE 15th
  International Symposium on Biomedical Imaging (ISBI 2018). pp. 1252--1255.
  IEEE (2018)

\bibitem{parisot2018disease}
Parisot, S., Ktena, S.I., Ferrante, E., Lee, M., Guerrero, R., Glocker, B.,
  Rueckert, D.: Disease prediction using graph convolutional networks:
  Application to autism spectrum disorder and alzheimer’s disease. Medical
  image analysis  (2018)

\bibitem{schaefer2017local}
Schaefer, A., Kong, R., Gordon, E.M., Laumann, T.O., Zuo, X.N., Holmes, A.J.,
  Eickhoff, S.B., Yeo, B.T.: Local-global parcellation of the human cerebral
  cortex from intrinsic functional connectivity mri. Cerebral Cortex
  \textbf{28}(9),  3095--3114 (2017)

\bibitem{vieira2017using}
Vieira, S., Pinaya, W.H., Mechelli, A.: Using deep learning to investigate the
  neuroimaging correlates of psychiatric and neurological disorders: Methods
  and applications. Neuroscience \& Biobehavioral Reviews  \textbf{74},  58--75
  (2017)

\bibitem{xingjian2015convolutional}
Xingjian, S., Chen, Z., Wang, H., Yeung, D.Y., Wong, W.K., Woo, W.c.:
  Convolutional lstm network: A machine learning approach for precipitation
  nowcasting. In: Advances in neural information processing systems. pp.
  802--810 (2015)

\end{thebibliography}






\end{document}